\documentclass{article}
\usepackage{ijcai19}
\usepackage{times}
\usepackage{url}
\usepackage[hidelinks]{hyperref}
\usepackage[utf8]{inputenc}
\usepackage[small]{caption}
\usepackage{graphicx}
\usepackage{amsmath}
\usepackage{amssymb}
\usepackage{booktabs}
\usepackage{algorithm}
\usepackage{algorithmic}
\usepackage{xcolor}
\title{Consensus Clustering: An Embedding Perspective, Extension and Beyond}

\author{
    Hongfu Liu$^1$\and Zhiqiang Tao$^2$ \and Zhengming Ding$^3$\\
    \small{\textnormal{$^1$Department of Computer Science, Brandeis University, $^2$Department of ECE, Northeastern University\\ $^3$Department of Computer, Information and Technology, IUPUI}}\\
    \textnormal{hongfuliu@brandeis.edu, zqtao@ece.neu.edu, zd2@iu.edu}
}

\usepackage{array}
\newcolumntype{L}[1]{>{\raggedright\let\newline\\\arraybackslash\hspace{0pt}}m{#1}}
\newcolumntype{C}[1]{>{\centering\let\newline\\\arraybackslash\hspace{0pt}}m{#1}}
\newcolumntype{R}[1]{>{\raggedleft\let\newline\\\arraybackslash\hspace{0pt}}m{#1}}

\begin{document}

\maketitle

\begin{abstract}
% what about a robust one?
Consensus clustering fuses diverse basic partitions (\emph{i.e.}, clustering results obtained from conventional clustering methods) into an integrated one, which has attracted increasing attention in both academic and industrial areas due to its robust and effective performance. Tremendous research efforts have been made to thrive this domain in terms of algorithms and applications. Although there are some survey papers to summarize the existing literature, they neglect to explore the underlying connection among different categories. Differently, in this paper we aim to provide an embedding prospective to illustrate the consensus mechanism, which transfers categorical basic partitions to other representations (\emph{e.g. }, binary coding, spectral embedding, etc) for the clustering purpose. To this end, we not only unify two major categories of consensus clustering, but also build an intuitive connection between consensus clustering and graph embedding. Moreover, we elaborate several extensions of classical consensus clustering from different settings and problems. Beyond this, we demonstrate how to leverage consensus clustering to address other tasks, such as constrained clustering, domain adaptation, feature selection, and outlier detection. Finally, we conclude this survey with future work in terms of interpretability, learnability and theoretical analysis. 
\end{abstract}
%with or without explicit objective function
%are also demonstrated within the consensus clustering framework

\section{Introduction}
Cluster analysis aims to separate a bunch of data points into different groups where the data points in the same group are similar to each other. As a fundamental research problem in machine learning area, tremendous efforts have been devoted to thrive this area in terms of advanced algorithms and diverse applications. Based on different assumptions on how the clusters form, various algorithms have been proposed, such as prototype K-means, density-based DBSCAN, connectivity-based hierarchical clustering, subspace clustering, etc. However, due to the unsupervised nature, it is difficult to select the most effective clustering methods and choose proper parameters among real-world applications. In light of this, consensus clustering has been proposed for pursing a robust partition.

Consensus clustering, also known as ensemble clustering, targets to integrate several diverse partition results from traditional clustering methods into a consensus one~\cite{Strehl02JMLR}. It has been widely recognized of robustness, consistency, novelty and stability over traditional clustering methods, especially in generating robust partitions, discovering novel structures, handling noisy features, and integrating solutions from multiple sources. The process of consensus clustering generally has two steps: basic partitions generation and consensus fusion. Given basic partitions as input, consensus clustering is in essence a fusion problem rather than a partitioning problem, which seeks for an optimal combinatorial result from basic partitions.
%generates a new partition from basic ones. 

%with different objective functions
Over the past years, many clustering ensemble techniques have been proposed, resulting in various of ways to face the problem together with new fields of application for these techniques. Generally speaking, consensus clustering can be divided into two categories, \emph{i.e.}, those with or without an explicit global objective function. The methods that do not set objective functions make use of some heuristics or meta-heuristics to find approximate solutions. Representative methods include co-association matrix-based~\cite{Fred05TPAMI,Lourenco13ML}, graph-based~\cite{Strehl02JMLR,Fern04ICML}, relabeling and voting based~\cite{Ayad08TPAMI} and locally adaptive cluster-based algorithms~\cite{Domeniconi09TKDD}. On another hand, the methods with explicit objectives employ global objective functions to measure the similarity between basic partitions and the consensus one. Representative solutions include K-means-like algorithm~\cite{Topchy03ICDM}, NMF~\cite{Li07ICDM}, EM algorithm~\cite{Topchy04SDM}, simulated annealing~\cite{Lu08AAAI} and combination regularization~\cite{Xie14KDD}.

%Representative methods include co-association matrix-based methods~\cite{Fred05TPAMI,Lourenco13ML}, graph-based algorithms~\cite{Strehl02JMLR,Fern04ICML}, relabeling and voting methods~\cite{Ayad08TPAMI} and locally adaptive cluster-based methods~\cite{Domeniconi09TKDD}. Methods with explicit objectives employ global objective functions to measure the similarity between basic partitions and the consensus one. Representative solutions include K-means-like algorithm~\cite{Topchy03ICDM}, NMF~\cite{Li07ICDM}, EM algorithm~\cite{Topchy04SDM}, simulated annealing~\cite{Lu08AAAI} and combination regularization~\cite{Xie14KDD}.

%Although they try to summarize as many methods as possible, they fail to uncover the deep connections among these methods. %Inspired by consensus clustering, some unsupervised tasks, such as constrained clustering, domain adaptation, feature selection, outlier detection could be solved in the consensus clustering framework.
%Although they aim to present a comprehensive introduction to existing methods, several key methods are not included and they neglect to uncover the deep connection among these methods. 

Considering the large number of consensus clustering methods, there are some excellent survey papers on this topic \cite{Sandro2011IJPRAI,boongoen2018cluster}. Although they aim to present a comprehensive introduction to existing methods, they may overlook several key methods and neglect to uncover the deep connection among these methods. In light of this, we provide an embedding prospective to understand the consensus mechanism in this study, which transfers the categorical basic partitions to other space for clustering, and unifies the above two major categories with or without explicit objective function in the literature. Specifically, several representative embedding strategies, such as binary coding, spectral decomposition, marginalized denoising, and random representations, are fully discussed for consensus clustering. Moreover, this paper also includes several variants of consensus clustering, \emph{e.g.}, fuzzy consensus clustering, multi-view clustering, and co-clustering. Beyond these extensions, we show that some important unsupervised tasks could be solved within a consensus clustering framework, such as constrained clustering, domain adaptation, feature selection, and outlier detection. 

This survey is organized as follows. We first give the preliminary knowledge in Section~\ref{sec:pd}. Then, we introduce several representative consensus clustering algorithms and link them from an embedding perspective in Section~\ref{sec:embedding}. Extensive extensions and applications of consensus clustering are provided in Section~\ref{sec:extension} and Section~\ref{sec:application}, respectively. Finally, we elaborate the future work with conclusive marks in Section~\ref{sec:conclusion}.

\section{Problem Definition}\label{sec:pd}
Let $\mathcal{X}=\{x_1,x_2,\cdots,x_n\}$ be a set of $n$ data points belonging to $K$ crisp clusters, denoted as $\mathcal{C} = \{C_1,\cdots,C_k\}$, where $C_k\bigcap C_{k'}=\emptyset,~\forall k\neq k'$, and $\bigcup_{k=1}^K C_k=\mathcal{X}$. Given $r$ basic partitions represented as $\Pi=\{\pi_1,\pi_2,\cdots,\pi_r\}$, each of which partitions $\mathcal{X}$ into $K_i$ clusters, and maps each data point to a cluster label ranged from $1$ to $K_i$. The goal of consensus clustering is to find an optimal consensus partition $\pi$ based on the input basic partitions $\Pi$ such as
\begin{equation}
    \Gamma(\Pi) \to \pi,
\end{equation}
where $\Gamma$ is a consensus function. 

The inputs of consensus clustering are several basic partitions $\Pi$, instead of the data matrix $\mathcal{X}$. Since clustering is an unsupervised task, the quality of basic partitions is difficult to assess in advance; and therefore, the diversity of basic partitions is taken into account instead. Generally, there are three generation strategies of basic partitions as follows.
%In the following, we will introduce three widely-used generation strategies of basic partitions
\begin{itemize}
\item \textit{Random Parameter Selection.} Almost every clustering method needs some parameters. Random Parameter Selection (RPS) randomly picks up the parameters in a certain range for the basic clustering method to generate several different basic partitions.
\item \textit{Random Feature Selection.} Different from RPS, Random Feature Selection (RFS) applies the same clustering method on a sub data derived by randomly selecting partial features. Note that the parameters within the basic clustering methods are fixed.
\item \textit{Different Methods Combination.} The last strategy is to generate a diverse basic partition set via different clustering methods.
\end{itemize}

%Since consensus clustering needs extra time compared with traditional clustering methods, usually a fast basic clustering method, such as $K$-means is chosen for its easy implementation and high efficiency. 

As generating basic partitions needs extra time compared with traditional methods, a fast basic clustering algorithm, such as K-means, is usually chosen due to its easy implementation and high efficiency. For the number of basic partitions, the theoretic analysis in \cite{luo2011consensus} and the experimental analysis in \cite{liu2016infinite} demonstrate that the performance of consensus clustering goes up and the variance becomes narrow with the increasing number of basic partitions. Hence, to achieve robustness, 100 basic partitions are empirically good enough for consensus clustering.

\section{An Embedding Perspective}\label{sec:embedding}
From the definition given in Section~\ref{sec:pd}, the input for consensus clustering is a set of basic partitions, rather than the original data matrix. However, although consensus clustering is a fusion problem in essence, due to the partition nature of its output, it is usually formulated as a classical partitioning problem upon the tool of co-association matrix~\cite{Fred05TPAMI}, which is defined as
\begin{small}
\begin{equation}\label{eq:s}
S(x,y)=\sum_{i=1}^r \delta(\pi_i(x),\pi_i(y)),~\delta(a,b)=\left\{
\begin{array}{ll}
1,&\textrm{if}~a=b\\
0,&\textrm{if}~a\neq b\\
\end{array}
\right..
\end{equation}
\end{small}The co-association matrix not only summarizes the categorical information of basic partitions into a pair-wise relationship, but also provides a chance to transform consensus clustering into a graph partitioning problem. Specifically, the co-association counts the co-occurrence of a pair of instances within the same cluster among $r$ basic partitions, and further measures the similarity between each pair of instances. By this means, the consensus clustering can be redefined as the classical graph partitioning problem. For example, \cite{Fred05TPAMI} applies the agglomerative hierarchical clustering on the co-association matrix for the final solution.

%Although there are several excellent surveys on consensus clustering~\cite{Sandro2011IJPRAI,boongoen2018cluster}, they mainly group the literature based on the methods, but fail to uncover the connection among different categories. Moreover, the hidden connection between the two major categories, the utility function based and co-association matrix based categories are not fully uncovered. In this paper, we aim to unified these two major categories and interpret the consensus clustering based on the co-association matrix from the embedding perspective with the following framework:

In this section, we focus on the co-association matrix and provide an embedding perspective to decompose the co-association for consensus clustering. To this end, some consensus clustering methods with or without explicit objective function are connected; moreover, it paves a path from consensus clustering to graph embedding, which leads to generalized tasks. Specifically, we have the
following framework to solve consensus clustering with embedding process: %following framework of consensus clustering from the embedding perspective: 
\begin{equation}
    \Gamma(\Pi) \to S \to Q \to \pi,
\end{equation}
where $Q$ is a representation matrix decomposed from $S$, and $\pi$ can be achieved by running some clustering algorithms on $Q$. It is worth noting that $S$ can be regarded as an $n$$\times$$n$ graph and $Q$ is an $n$$\times$$d$ matrix, where $d$ denotes the dimension of the embedded space. The process of $\Gamma(\Pi) \to S$ is to summarize the information from basic partitions, while the embedding of $S \to Q$ gives opportunities to understand consensus clustering from binary coding, spectral decomposition, marginalized denoising, random representation and other aspects. 
% an nxn matrix or a nxn matrix?

\subsection{Binary Coding}
%Binary coding is the simplest way to embed the co-association matrix into a sparse matrix with the belonging information, which can be even directly achieved from $\Pi$. 
Binary coding provides a simple way to embed co-association matrix into sparse representations, which could be directly achieved from $\Pi$. 
Let $B=\{b_l|1$$\leq$$l$$\leq$$n\}$ be a binary data set derived from the set of $r$ basic partitions $\Pi$ as follows:
\begin{equation}\label{eq:binary}
\begin{split}
&b_l=\langle b_{l,1},\cdots,b_{l,i},\cdots,b_{l,r}\rangle,~\textrm{with}\\
&b_{l,i}=\langle b_{l,i1},\cdots,b_{l,ij},\cdots,b_{l,iKi}\rangle,~\textrm{and}\\
&b_{l,ij}=\left\{
  \begin{array}{ll}
    1,&\textrm{if}~\pi_i(x_l)=j\\
    0,&\textrm{otherwise}
  \end{array}
\right.,
\end{split}
\end{equation}

%\vspace{-0.3cm}
%\scriptsize

\begin{table*}[t]
  \caption{Sample KCC Utility Functions}\label{tab:Uexample}\vspace{-0.2cm}
  \centering
  \scalebox{0.86}{
  %\begin{tabular}{L{1.5cm}L{4.5cm}L{4.5cm}L{4.5cm}}
  \begin{tabular}{L{1.2cm}lll}
    \toprule
     & \multicolumn{1}{l}{$\mu(m_{k,i})$} & \multicolumn{1}{l}{$U_\mu(\pi,\pi_i)$} & \multicolumn{1}{l}{$f(b_{l},m_k)$} \\
    \midrule
    $U_c$ & $\|m_{k,i}\|_2^2-\|P^{(i)}\|_2^2$ & $\sum_{k=1}^K p_{k+}\|m_{k,i}\|_2^2 - \|P^{(i)}\|_2^2$ &  $\sum_{i=1}^r  \|b_{l,i}-m_{k,i}\|_2^2$ \\
    %\hline
    $U_H$ & $(-H(m_{k,i}))-(-H(P^{(i)}))$ & $\sum_{k=1}^K p_{k+}(-H(m_{k,i})) - (-H(P^{(i)}))$ & $\sum_{i=1}^rD(b_{l,i}\|m_{k,i})$ \\
    %\hline
    $U_{\cos}$ & $\|m_{k,i}\|_2-\|P^{(i)}\|_2$ & $\sum_{k=1}^K p_{k+}\|m_{k,i}\|_2 - \|P^{(i)}\|_2$ & $\sum_{i=1}^r (1-\cos(b_{l,i},m_{k,i}))$ \\
    %\hline
    $U_{L_p}$ & $\|m_{k,i}\|_p-\|P^{(i)}\|_p$ & $\sum_{k=1}^K p_{k+}\|m_{k,i}\|_p - \|P^{(i)}\|_p$ & $\sum_{i=1}^r  (1-\sum_{j=1}^{K_i}b_{l,ij}(m_{k,ij})^{p-1}/\|m_{k,i}\|_p^{p-1})$\\
    \bottomrule
    %\multicolumn{4}{l}{Notes: (1) $P^{(i)}$ -- $\langle p_{+1}^{(i)}, \cdots, p_{+j}^{(i)}, \cdots, p_{+K_i}^{(i)} \rangle$; (2) $P_k^{(i)}$ -- $\langle p_{k1}^{(i)}, \cdots, p_{kj}^{(i)}, \cdots, p_{kK_i}^{(i)} \rangle$; }\\
    \multicolumn{4}{l}{Note: $P^{(i)}$ is the cluster size portion of $\pi_i$, cos is the cosine distance, $H$ and $D$ denote the Shannon entropy and KL-divergence, respectively.}\vspace{-5mm}
  \end{tabular}}
\end{table*}

\noindent where ``$\langle~\rangle$'' indicates a transversal vector. As shown in Eq.~(\ref{eq:binary}), the binary coding is nothing but concentrating all the 1-of-$K_i$ coding of basic partitions, where $K_i$ is the cluster number for $\pi_i$. Therefore, for binary coding, the embedded representation $Q=B$ is an $n$$\times$$\sum_{i=1}^rK_i$ binary data matrix with $|q_{l,i}|=1,~\forall~l,i$. Moreover, by taking a close look at $B$ in Eq.~\eqref{eq:binary} and $S$ in Eq.~\eqref{eq:s}, we may conclude that $S=BB^\top$, which is a matrix-wise formulation of Eq.~\eqref{eq:s}. Thus, although the binary coding $Q$ in Eq.~\eqref{eq:binary} is directly obtained from the basic partitions, it can also be regarded as the sparse embedding from $S$ without any information loss.

The binary coding also provides an efficient solution via K-means-based Consensus Clustering (KCC)~\cite{Wu15TKDE}, which transforms the utility function based consensus clustering into a simple K-means clustering. Different from the co-association matrix based methods focusing on the pair-wise similarity between instances, the utility function based methods measure the similarity on a partition level and formulate the consensus clustering as a utility maximization problem. In the scenario of KCC, we have 
\begin{equation}\label{eq:kcc}
\min_{\pi} \sum_{k=1}^K\sum_{b_l\in C_k}\sum_{i=1}^r f(b_{l,i}, m_{k,i})  \Leftrightarrow \max_{\pi} \sum_{i=1}^r U(\pi,\pi_i)\text{,}
\end{equation}
where $f$ is the K-means distance~\cite{Wu12TFS}, $m_{l,i}$ the $k$-th centroid of $i$-th basic partition, and $U$ the corresponding KCC utility function~\cite{Wu15TKDE}. Benefiting from Eq.~\eqref{eq:kcc}, the consensus partition is achieved by running K-means on binary codes with a nice utility interpretation. Based on the binary coding, they established a necessary and sufficient condition for KCC utility functions from a continuously differentiable convex function, and further built the link between K-means distance and certain type of utility functions. Table~\ref{tab:Uexample} shows some core continuously differentiable convex functions $\mu$, utility functions $U_{\mu}$ and the corresponding K-means distances $f$ on with binary coding. For example, if we choose the classical squared Euclidean distance for K-means, the corresponding utility function is the well-known categorical utility function~\cite{Mirkin01ML}. Moreover, some new types of utility functions can be designed based on $L_p$ norm and entropy for text clustering~\cite{liu2015dias} and patient stratification~\cite{liu2017entropy}. The utility function family provides powerful tools to extend consensus clustering in other tasks, which will be discussed in Section 5. 

\subsection{Spectral Decomposition}
The co-association matrix can be treated as a similarity graph, which leads the consensus clustering to a graph partitioning problem. Besides agglomerative hierarchical clustering, spectral clustering is another widely-used graph partitioning algorithm. It is naturally to conduct spectral decomposition on the co-association matrix for low-dimension embedding. However, its time complexity is expensive due to the singular vector decomposition. To tackle the algorithmic cost, Liu et al. proposed Spectral Ensemble Clustering (SEC) and uncovered the equivalent relationship between the spectral decomposition and weighted K-means~\cite{Liu15KDD,liu2017spectral} as follows:
\begin{equation}\label{eq:spectral}
\begin{split}
&\max_{Z} \textup{tr}(Z^\top D^{-1/2}{S}{D}^{-1/2}{Z}) \\
\Leftrightarrow  &\min_{\pi} \sum_{k=1}^K\sum_{b_l\in C_k}\sum_{i=1}^r w_lf(b_{l,i}/w_l, m_{k,i}),
\end{split}
\end{equation}
where $m_{k,i} = \frac{\sum_{b_l\in C_k} b_{l,i}}{\sum_{b_l\in C_k} w_l}$, and $w_l=D(x_l,x_l)$. The left part in Eq.~\eqref{eq:spectral} is similar to the classical normalized-cut spectral clustering, yet applies co-association matrix to replace the original Laplacian graph, where ${D}$ is a diagonal degree matrix with ${D}_{ll}=\sum_{j}{S}_{jl}, 1$$\leq$$j, l$$\leq$$n$, and ${Z}$ is a scaled indicator matrix. A well-known solution to SEC is to run K-means on the top $K$ largest eigenvectors of ${D}^{-1/2}{SD}^{-1/2}$ for obtaining the final consensus partition $\pi$~\cite{Yu03ICCV}, which consists of $K$ clusters $C_1, C_2, \cdots, C_K$.

%To address this challenge, one feasible way is to find a more efficient yet equivalent solution for SEC with elegant embedding. Different from seeking a dense low-dimension decomposition via singular vector decomposition, a sparse high-dimension spectral embedding can be elegantly designed to achieve the same task in the right side of Eq.~\eqref{eq:spectral}. 

Performing spectral clustering on the co-association matrix, however, suffers from huge time complexity originated from both building the matrix and conducting the clustering. 
To address this challenge, one feasible solution is to find an efficient yet equivalent embedding method to solve SEC. Different from seeking for dense low-dimension representations via singular vector decomposition, a sparse high-dimension spectral embedding can be neatly designed to achieve the same goal with the right side of Eq.~\eqref{eq:spectral}. 
The embedding matrix $Q=\{b_{l}/w_l\}$ is applied for weighted K-means, where $b_{l,i}$ is defined in Eq.~\eqref{eq:binary} and $w_l$ lies in ${D}$ can also be directly obtained  from basic partitions as $w_l=\sum_{i=1}^r\sum_{l=1}^{n}\delta(\pi_i(x),\pi_i(x_l))$.

SEC explicitly transforms the consensus clustering based on spectral decomposition into a weighted K-means clustering, where the transformation dramatically reduces the time and space complexities from $\mathcal{O}(n^3)$ and $\mathcal{O}(n^2)$, respectively, to roughly $\mathcal{O}(n)$. Similar to the binary coding in Eq.~\eqref{eq:binary}, the spectral decomposition $Q$ is of high sparsity for efficient memory, where $r$ elements in each row are non-zeros. 

Along this line, robust spectral ensemble clustering imposes a low-rank constraint on the co-association matrix for delivering robust embedding representations~\cite{tao2016robust,tao2019robust}; whilst Huang et al. considered the co-association matrix with uncertainty and proposed locally weighted graph partitioning for the consensus solution~\cite{huang2015combining,huang2018locally}. 
%Beyond the standard co-association matrix, Huang et al. considered the co-association matrix with uncertainty and proposed locally weighted graph partitioning for the consensus solution~\cite{huang2015combining,huang2018locally}. 

%The constant 1 is added to the last column of $B$ and corrupt it with $s$ level drop-out noise. 
\subsection{Marginalized Denoising}
The above two strategies employ the fixed co-association matrix with limited basic partitions for embedding. With the increasing number of basic partitions, consensus clustering achieves better performance and lower variance~\cite{luo2011consensus}. To tackle the number of basic partition, Infinite Ensemble Clustering (IEC) aims to fuse infinite basic partitions for robust embedding representations~\cite{liu2016infinite}. Rather than directly obtaining the expectation of a co-association matrix, IEC purses the expectation representation of basic partition, \emph{i.e.}, $Q=\mathbb{E}[B]$. 
%The key idea of IEC is that extra incomplete basic partitions are generated by removing the labels and this process is repeated \emph{infinite} times. %这句话觉得好奇怪
To achieve this, Denoising Auto-Encoder~\cite{Chen12ICML} with dropout noises is employed to learn the hidden representation for $\mathbb{E}[B]$. For linear denoising Auto-Encoder, the corresponding mapping for $W$ between input and hidden representations is calculated by
\begin{equation}\label{eq:W}
W = \mathbb{E}[U]\mathbb{E}[V]^{-1},
\end{equation}
where $U = r{S}$ and $V = B^\textup{T}B = {\Sigma}$. We corrupt $B$ with $s$ level drop-out noises and add the constant 1 to its last column. 
Let $v = [1-s,\cdots , 1-s, 1]$, we have $\mathbb{E}[{U}]_{ij} = {\Sigma}_{ij}v_j$ and $\mathbb{E}[V]_{ij} = {\Sigma}_{ij}v_i \tau (i,j,v_j)$. Here $\tau (i,j,v_j)$ returns 1 with $i=j$, and returns $v_j$ with $i \neq j$. After getting the mapping matrix, $Q=BW^\textup{T}$ is used as the marginalized representation for consensus clustering. 

%The non-linear version of IEC is provided in \cite{liu2018infinite}. 
The marginalized denoising embedding incorporates extra incomplete basic partitions for robust representation, which embeds \emph{infinite} basic partitions into a fixed dimension via neural networks. Thanks to this marginalized embedding, a method in deep learning domain can be employed to handle the challenging problem in consensus clustering area. 

\subsection{Random Representation}

Elements in a co-association matrix might be differently treated according to their importance and uncertainty. In SEC, the points in large clusters are assigned with large weights to resist the negative impact from noises. However, such weighting strategy only considers the information from independent basic partitions, yet neglects the graph structure of co-association matrix. To further capture the higher-order information embedded in the co-association matrix, Probability Trajectories employ the random walks to explore the graph and propose to derive a dense pair-wise similarity measure based on the probability trajectories of random walkers. The random walk process driven by a new transition probability matrix is utilized to explore the global information in the graph~\cite{huang2016robust}. Probability Trajectory based similarity and accumulation with hierarchical clustering are calculated for the consensus clustering.%啥叫accumulation with hierarchical clustering？

\subsection{Others Embedding}
Graph embedding is an increasing hot topic in network analysis, and it fits the consensus clustering task well as will be showcased in Section~\ref{sec:application}. Generally, factorization, random walk and deep learning are three major categories. Among existing works, Locally linear embedding, Laplacian Eigenmaps, DeepWalk, LINE, node2vec, graph convolutional neural networks and variational graph auto-encoders are some representative methods, which can be easily adapted to consensus clustering with co-association matrix. More details on graph embedding can be refer to \cite{goyal2018graph}.

In summary, the co-association matrix is naturally a pair-wise similarity graph, which directly measures the probability of a pair of instances in the same cluster, rather than the distance. From the graph view, many conventional embedding methods can be directly applied for consensus clustering. Moreover, some novel ways mentioned above are particularly designed to learn embedding representations according to the property of co-association matrix. 

\section{Extension of Consensus Clustering}\label{sec:extension}
In this section, we elaborate some extensions of consensus clustering from different settings, problems and applications. 

\textbf{Fuzzy Consensus Clustering.} Fuzzy clustering has been widely used in many real-world application domains for a long time, which provides a soft assignment for the degree membership in each cluster. The introduction of consensus clustering to fuzzy clustering becomes natural, and fuzzy consensus clustering thus emerges as a new research frontier. Similar to KCC framework, Wu et al. proposed FCC utility functions that can transform FCC to a weighted piece-wise fuzzy c-means clustering problem~\cite{wu2017fuzzy}. 
%Su et al. employed a fuzzy hierarchical graph to represent the relationships between the resulting base clusters, which applies fuzzy c-means and hierarchical clustering in generating base cluster and implementing consensus function respectively~\cite{su2015hierarchical}.

%\textbf{Hierarchical Consensus Clustering.} Different from partitional or flat clustering, hierarchical clustering results are often more complex, which are typically represented as dendrograms or trees. Hierarchical Ensemble Clustering~\cite{zheng2014framework} takes as input both partitional and hierarchical clustering, learns a ultra-metric distance from the aggregated distance matrices, and generates final hierarchical clustering. 

\textbf{Simultaneous Clustering and Ensemble}. The standard consensus clustering only takes basic partitions as input, rather than data matrix. However, these co-association matrices summarized from these basic partitions may suffer from information loss when computing the similarity between data points, because they only utilize the categorical data provided by multiple basic partitions, yet neglect rich information from raw features. This problem can badly undermine the underlying cluster structure in the original feature space, and thus degrade the clustering performance. Tao et al. proposed Simultaneous Clustering and Ensemble (SCE) to alleviate such detrimental effect, which employs a similarity matrix from raw features to enhance the co-association matrix summarized by multiple basic partitions~\cite{tao2017simultaneous}. Two neat closed-form solutions given by eigenvalue decomposition are provided for SCE.

\textbf{Multi-View Clustering.} Multi-view data are extensively accessible nowadays thanks to various types of features, viewpoints and sensors~\cite{zhao2017multi}. Among the literature, late fusion is a representative category for multi-view clustering, which generates basic partitions first and applies consensus clustering to fuse them together. Along this line, Tao et al. employed the low-rank and sparse decomposition on the co-association matrix to explicitly consider the connection between different views and handle the noises in each single view~\cite{tao2017ensemble}. Liu and Fu incorporated the generation of multi-view basic partitions and the fusion of consensus clustering in an interactive way, \emph{i.e.}, the consensus clustering guides the generation of basic partitions, and high-quality basic partitions positively contribute to the consensus clustering in return~\cite{liu2018consensus}.

%and has been successfully applied to text mining, bioinformatics, chemometrics and recommendation systems, lexicon and concept construction in natural language processing etc.
\textbf{Co-Clustering.} Co-clustering aims to simultaneously partition the instances and features to obtain co-clusters. Due to the ability to capture the joint cohesion among instances and features, co-clustering outperforms the traditional clustering methods. However, co-clustering still suffers from the divergences of different methods. It is natural to tackle this challenge with a consensus fashion. Spectral Co-Clustering Ensemble (SCCE) has been presented to perform ensemble tasks on basic row/column clusters of a dataset simultaneously, and obtains an optimization co-clustering result~\cite{huang2015spectral}. SCCE is a matrix decomposition based approach which can be formulated as a bipartite graph partitioning problem and solved efficiently with the selected eigenvectors. 

\textbf{Big Data Clustering.} Although consensus clustering outperforms traditional clustering algorithms, the computational cost is a downside to some extent. 
When it comes to ``big data'', consensus clustering seems not a good choice at the first glance. This is mainly due to the fact that the consensus fusion (\emph{i.e.}, late fusion) requires to generate diverse and enough basic partitions in advance, which is quite time consuming even with an efficient clustering algorithm on large-scale datasets.
To handle this challenge, incomplete basic partitions are taken into consideration, where the big data are split into several small subsets via row and column random sampling, then incomplete basic partitions are obtained from these subsets, finally consensus clustering is called to fuse these incomplete basic partition together. Thanks to the robustness of consensus clustering, the performance of consensus clustering with incomplete basic partitions does not suffer from a significant drop.

%When it comes to big data clustering, consensus clustering seems not a good choice even with an efficient algorithm at the first glance since diverse and enough basic partitions are generated first for later fusion. 

%\textbf{Applications.}
%Since consensus clustering has the same task with the classical cluster analysis, it also enjoys many real world applications, such as document clustering, patient stratification, biological analysis, image segmentation or co-segmentation, recommendation system, weather analysis, social media, network analysis, chemical structures prediction, internet security, phishing profiling, and so on. 

\section{Beyond Consensus Clustering}\label{sec:application}
In this section, we illustrate several research problems related to consensus clustering, which employs the utility function or co-association as a regularizer for a specific learning task. 

%Inspired by consensus clustering, partition level constrained clustering aims to find a partition which captures the intrinsic structure from the data itself, and also agrees with the partition level side information as %follows:
\textbf{Constrained Clustering.} Constrained clustering applies side information to guide the clustering process. Since clustering is an orderless partition, pairwise constraints are traditionally employed to further improve the performance of clustering. Specifically, Must-Link and Cannot-Link constraints represent that two instances should lie in the same cluster or not, respectively. Within the framework of pairwise constraints, we avoid answering the mapping relationship among different clusters and at the first thought it is easy to make the Must-Link or Cannot-Link decision for pairwise constraints. However, it is highly risky to make the pairwise decision without any prior knowledge or references. In contrast to traditional pairwise constraint, partition level side information or partial labels are incorporated in constrained clustering~\cite{liu2015clustering}. 
Specifically, a partition-level constrained clustering, which captures the intrinsic structure from data itself and also agrees with the partition level side information, is formulated as %follows:
\begin{equation}\label{eq:constrained}
    \min_{H}\mathcal{J}(X;H)-\lambda U(H\circ M,P),
\end{equation}
where $\mathcal{J}$ is a clustering loss with $X$ and $H$ denoting the data matrix and assignment matrix, $P$ is the partial partition level side information, $M$ is a binary matrix to indicate whether a data point has side information. In Eq.~\eqref{eq:constrained}, the utility function treats the side information as a whole and plays a role as a regularizer in making the final partition close to the partial labels as much as possible. 

\textbf{Domain Adaptation.} Unsupervised domain adaptation aims to employ the auxiliary labeled source data to predict labels of another related unlabeled target data. Since the distributions of source and target data are different, the alignment and transfer tasks are two key challenges in domain adaptation, where a projection matrix aligns the source data and target data in a common space. Assuming that the projection matrix is given and the source data have labels, the transfer task can be formulated the domain adaptation problem as a partition as follows:% 
\begin{equation}\label{eq:adaptation}
    \min_{H_S,H_T}\mathcal{J}(Z_S,Z_T;H_S,H_T)-\lambda U(H_S,Y_S),
\end{equation}
where $Z_S$ and $Z_T$ are the source and target data representation in the common space after distribution alignment, $H_S$ and $H_T$ are the learnt label indicator matrices for source and target data. Similarly, the utility function preserves the complete source structure to guide the target data clustering.  

\textbf{Feature Selection.} In the unsupervised feature selection, the pseudo labels generated by some clustering method are employed as a criteria to guide feature selection. Consensus Guided Unsupervised Feature Selection (CGUFS) introduces consensus clustering to generate pseudo labels for feature selection~\cite{liu2018feature}. 
Specifically, fusing multiple basic partitions helps reduce the risk of noisy features and provide the robust consensus labels to guide feature selection process:
\begin{equation}
    \min_{H,Z,G} \alpha \mathcal{J}(\Pi, H) + \|XZ-HG\|^2_{\textup{F}} + \beta\|Z\|_{2,1},
\end{equation}
where $J$ is the consensus clustering loss, $H$ is the consensus partition in the matrix formulation, $Z$ is the mapping matrix for feature selection, and $G$ is an alignment matrix. The above framework jointly learns the consensus partition and feature selection matrix. To this end, CGUFS not only conducts robust feature selection, but also provides the interpretability of consensus clustering. 

\textbf{Outlier Detection.} Cluster analysis and outlier detection are closely coupled tasks, \emph{i.e.}, clustering structure can be easily discovered if we are able to identify a few outliers; at the same time, the outliers can be identified if the clustering structure is discovered, since outliers belong to none of clusters. Compared with conducting the joint clustering and outlier detection in the original feature space, the partition space is more appealing. The benefits to transform the original space into the partition space lie in (1) the binary value indicates the cluster-belonging information, which is particularly designed according to the definition of outliers, and (2) compared with the continuous space, the binary space is much easier to identify the outliers due to categorical features. 
Clustering with Outlier Removal (COR) employs the Holoentropy on the partition space for joint clustering and outlier detection~\cite{liu2018clustering}, which is completely solved via K-means by
%via a K-means{-}{-} as follows:
\begin{equation}
\min_{C,O} \sum_{k=1}^Kp_k HL(C_k) \Leftrightarrow \min_{C,O} \sum_{k=1}^K\sum_{\tilde{b_l}\in C_k} \sum_{i=1}^rD(\tilde{b_l},\tilde{m_k}),
\end{equation}
where ${C,O}$ denote $K$ clusters and the outlier set, $HL$ and $D$ are the Holoentropy and KL-divergence, $\tilde{B}=\{\tilde{b_l}\}$ and $m_k$ is the $k$-th centroid of $\tilde{B}$. $\tilde{B}$ can be obtained via generating basic partitions on the original data with $\tilde{B} = [B\ \overline{B}]$, where $B$ is the binary coding in Eq.~\eqref{eq:binary} and $\overline{B}$ is a flip matrix of $B$ with zeros being ones and ones being zeros. The partition space derived from basic partitions enables COR not only to identify outliers, but also to fuse basic partition to achieve consensus clustering. From this view, Holoentropy can be regarded as the utility function to measure the similarity between the basic partitions and the final one.

\textbf{Consensus Graph Embedding.} Graph embedding techniques have been well studied in recent years, which target to learn a low-dimensional embedding space with graph structure to facilitate downstream tasks. Besides using the real graph structured data, many graph constructions methods are proposed to exploit the local relationship between data samples, such as $k$-nearest neighbor ($k$-NN) graph, $\varepsilon$-neighborhood graph, and b-matching graph. From this line, ensemble clustering can be also used to construct a robust graph. Particularly, consensus graph built on the co-association matrix (as shown in Eq.~(\ref{eq:s})) provides an important alternative input for graph embedding tasks, as it captures various cluster structures underlying in real-world data~\cite{Fred05TPAMI}. Moreover, by leveraging the recent graph convolutional networks (GCN)~\cite{KipfW16}, a deep consensus graph embedding model could be provided as the following. 

Let $X \in \mathbb{R}^{n \times d}$ be the feature matrix of $\mathcal{X}$, where each data point is represented by a $d$-dimension feature vector, then a single spectral graph convolutional layer with consensus graph $S$ is defined as:
\begin{equation}\label{eq:g_Z}
Z = \tilde{D}^{-\frac{1}{2}}\tilde{S}\tilde{D}^{-\frac{1}{2}}XW,
\end{equation}
where $\tilde{S} = I_{N} + S$, $\tilde{D} \in \mathbb{R}^{n \times n}$ is a diagonal matrix with $\tilde{D}_{ii}=\sum_{j}\tilde{S}_{ij}$, $W \in \mathbb{R}^{d \times m}$ represents the learnable network parameters, $m<d$ is the dimension of embedding space, and $Z \in \mathbb{R}^{n \times m}$ denotes the final embedding representations.

%could be defined with the following layer-wise propagation rule:
Based on Eq.~(\ref{eq:g_Z}), a multi-layer network is further provided in a layer-wise propagation fashion by
\begin{equation}\label{eq:GCN}
Z^{(l+1)} = g(Z^{(l)}, S) = 
\varphi(\tilde{D}^{-\frac{1}{2}}\tilde{S}\tilde{D}^{-\frac{1}{2}}Z^{(l)}W^{(l)}),
\end{equation}
where $g(Z^{(l)}, S)$ refers to the graph convolution network, $\varphi(\cdot)$ represents an activation function, such as the ReLU or Sigmoid function, and ${W}^{(l)} \in \mathbb{R}^{m_{l-1} \times m_l}$ ($m_0=d$) is the $l^{th}$ layer's filter parameters matrix. ${Z}^{(l)}$ denotes the graph embeddings given by the $l^{th}$ GCN layer, while ${Z}^{(0)}$ indicates the input feature matrix ${X}$. The consensus graph embedding network provides a learnable way to utilize the cluster structure provided by ensemble clustering, and could be trained upon a wide range of applications.

In summary, consensus clustering provides a chance to improve the cluster-based tasks, such as feature selection, outlier detection, where consensus clustering generates high-quality pseudo labels to guide the following task. Moreover, the co-association matrix plays as a robust and consensus graph, which can be alternatively used over traditional graphs, such as Lapalacian graph, $k$-nearest neighbor graph, etc.

\section{Conclusions and Future Work}\label{sec:conclusion}
In this paper, we provided a consensus clustering survey from an embedding perspective, where embedding representations are learnt from the co-association matrix and followed by a classical clustering algorithm to eventually achieve a consensus partition. Generally speaking, several representative embedding strategies, such as binary coding, spectral decomposition, marginalized denoising, random representation were fully discussed. Moreover, the variants of consensus cluster were extensively illustrated in terms of different settings, problems and application. Finally, inspired by consensus clustering, sevaral important unsupervised tasks, such as constrained clustering, domain adaptation, feature selection, and outlier detection, were introduced to solve within a consensus clustering framework. Beyond that, more efficient and effective consensus clustering algorithms in different scenarios, including incomplete data, heterogeneous data and large-scale data, are welcome to thrive this area. In the meanwhile, there are still several not adequately addressed open problems in this direction, which remain as future works in the following. 
%Beyond that more efficient and effective consensus clustering algorithms in different scenarios, such as incomplete data, heterogeneous data and large-scale data are welcome to thrive this area, there are still several not adequately addressed open problems in this direction, which remain as the future work:

\vfill
\textbf{Interpretable Consensus Clustering.} After achieving data partition, a natural question is to understand the uncovered cluster structure or interpret the meaning of clusters. Although consensus clustering outperforms the traditional clustering methods, it conducts on the co-association matrix or its derived embedding representations, rather than the original feature matrix. In such a case, it loses the power to interpret the uncovered clusters via some key original features. This poses the interpretable consensus clustering, which not only achieves a robust consensus partition result, but also interprets clusters with original or recovered semantics.

\vfill
\textbf{Learnable Consensus Clustering.} Currently the most consensus clustering algorithms take the fixed basic partitions as input, which means the basic partition generation and consensus fusion are two separated processes. In fact, basic partitions play a crucial role in the consensus clustering process. Hence, a learnable consensus clustering model which aims to jointly generate basic partitions and fuse them together in an end-to-end framework will be highly recommended. Particularly, the consensus clustering can further guide the high-quality basic partition generation; and high-quality basic partitions produce the consensus clustering in an interactive fashion. 
Especially with the rapid development of deep neural networks, such a framework might be achieved with drop-out or other mechanism to generate diverse basic partitions, which can also be assembled together in a fusion layer. 

\vfill
\textbf{Theoretical Analysis of Consensus Clustering.} The theoretical analyses of consensus clustering have not been fully explored yet. Although consensus clustering empirically achieves robust and high performance, the question that why consensus clustering would achieve robust performance should be urgently answered with theoretical solutions. The relationship between original features and partition space is also under-explored. Moreover, the generalization, error bound, convergence properties of consensus clustering algorithms are strongly encouraged to be solved in theory. Some techniques for supervised ensemble classification might be transferred on the analysis of consensus clustering.

\clearpage
\begin{small}
\bibliographystyle{named}
\bibliography{CC-Survey}
\end{small}
\end{document}